\documentclass{article}

\usepackage[preprint]{neurips_2025}

\usepackage[utf8]{inputenc} 
\usepackage[T1]{fontenc}    
\usepackage{hyperref}       
\usepackage{url}            
\usepackage{booktabs}       
\usepackage{amsfonts}       
\usepackage{nicefrac}       
\usepackage{microtype}      
\usepackage{xcolor}         
\usepackage{graphicx}
\usepackage{subfigure}
\usepackage{subcaption}
\usepackage{comment}
\usepackage{placeins}
\usepackage{amsmath}

\title{Concept Based Explanations and Class Contrasting}

\author{%
  Rudolf Herdt \\
  Center for Industrial Mathematics \\
  University of Bremen \\
  Germany \\
  \texttt{rherdt@uni-bremen.de} \\
  \And
  Daniel Otero Baguer \\
  Center for Industrial Mathematics \\
  University of Bremen \\
  Germany \\
}

\begin{document}

\maketitle

\begin{abstract}
Explaining deep neural networks is challenging, due to their large size and non-linearity. In this paper, we introduce a concept-based explanation method, in order to explain the prediction for an individual class, as well as contrasting any two classes, i.e. explain why the model predicts one class over the other. We test it on several openly available classification models trained on ImageNet1K. We perform both qualitative and quantitative tests. For example, for a ResNet50 model from pytorch model zoo, we can use the explanation for why the model predicts a class 'A' to automatically select four dataset crops where the model does not predict class 'A'. The model then predicts class 'A' again for the newly combined image in 91.1\% of the cases (works for 911 out of the 1000 classes).\footnote{The code including an .ipynb example is available on github: \\ https://github.com/rherdt185/concept-based-explanations-and-class-contrasting }
\end{abstract}

\section{Introduction}

Interpreting deep neural networks is challenging.
At the same time, the range of their use cases widens, also into more critical areas like medicine \cite{JANSEN2023161}.
As the stakes of the application become higher, interpretability of the behavior of the models becomes more important.

There is a wide range of existing work that focuses on the challenge of explaining such deep neural networks.
Earlier methods employ heat maps, where the idea is to highlight pixels in the input that are important for the prediction of the model.
Those are usually gradient based (like DeepLift \cite{shrikumar2019learning}, Integrated Gradients \cite{pmlr-v70-sundararajan17a}, Smooth Grad \cite{SmoothGrad} or GradCAM \cite{Selvaraju_2017_ICCV}), or pertubation based (like RISE \cite{petsiuk2018rise}).
There are some problems with such heatmap based methods.
Previous work raised concerns about the reliability of some gradient based methods \cite{SanityCheckSaliencyMaps} and they may also not increase human understanding of the model \cite{kim2018interpretabilityfeatureattributionquantitative}.
A more fundamental problem is while those heatmap methods show where the model is seeing something, they do not explain what the model is seeing there, which becomes a problem if what a human sees in the area highlighted by the heatmap does not align with what the model is seeing there \cite{Fel_2023_CVPR}.

As a consequence, other methods emerged that focus on concept based explanation.
\cite{olah2018the} combined saliency maps used in hidden layers with visualizing the internal activations at those layers (therefore combining where the model is seeing something, with what it is seeing there).
TCAV \cite{kim2018interpretabilityfeatureattributionquantitative} tests the importance of human defined concepts for the prediction of a specific class (e.g. how important the concept of stripes is for the prediction of the class zebra).
ACE \cite{ghorbani2019automaticconceptbasedexplanations} goes a step further, and extracts those concepts automatically from a dataset, removing the need for human defined concepts.
CRAFT \cite{Fel_2023_CVPR} takes this another step further, and allows to create concept based heatmaps explaining individual images.
Further CRAFT uses a human study to show the practical utility of the method.

\begin{figure}[h!]
\vskip 0.2in
\begin{center}
\centerline{\includegraphics[width=0.4\columnwidth]{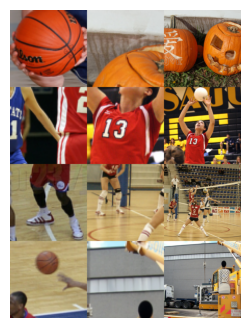}}
\caption{Example of using our method to explain why a ResNet50 model from pytorch model zoo predicts the class basketball. We extract the four main concepts for the class basketball (shown in the left column). The middle column shows the same concepts, but we only sample the crops from images where the model does not predict basketball for the crop and the whole image it came from (crucially to note, is that for the combined middle column image the model predicts basketball again with a confidence of 0.96). The four crops in the middle row were predicted as (from top to bottom): jack-o-lantern, volleyball, volleyball, horizontal bar. But for the model, the combined image from crops of them is predicted as basketball. This works not only for basketball, but for 911 out of the 1000 of the classes. The last column shows the original images that the middle column images were cropped out of. In this instance, even taking the second and third image of the middle row together gets to model to predict basketball with a confidence of 0.87, which points to a bias in the model, since a numbered jersey combined with feet on a hall floor should not be predicted as basketball.}
\label{fig:basketball}
\end{center}
\vskip -0.2in
\end{figure}

We also propose a concept based method, both for explaining a single class (like ACE or CRAFT), as well as contrasting any two classes.
Unlike ACE as well as CRAFT that both first extract the concepts, and later score them (ACE using TCAV and CRAFT using sobol indice), we first score the activations using attribution and afterwards extract the concepts (so we have no scoring after we extracted the concepts, we assume that all extracted concepts are relevant for the class).
And we get a new set of concepts per class.
Further, we automatically score the explanations using the model itself.
%
The idea for the automatic score is that concept based methods extract concepts that are important for the prediction of a given class, and show them to the user in the form of one or more dataset examples (e.g. crops of dataset images).
\begin{figure*}[tb]
\vskip 0.2in
\begin{center}
\centerline{\includegraphics[width=\columnwidth]{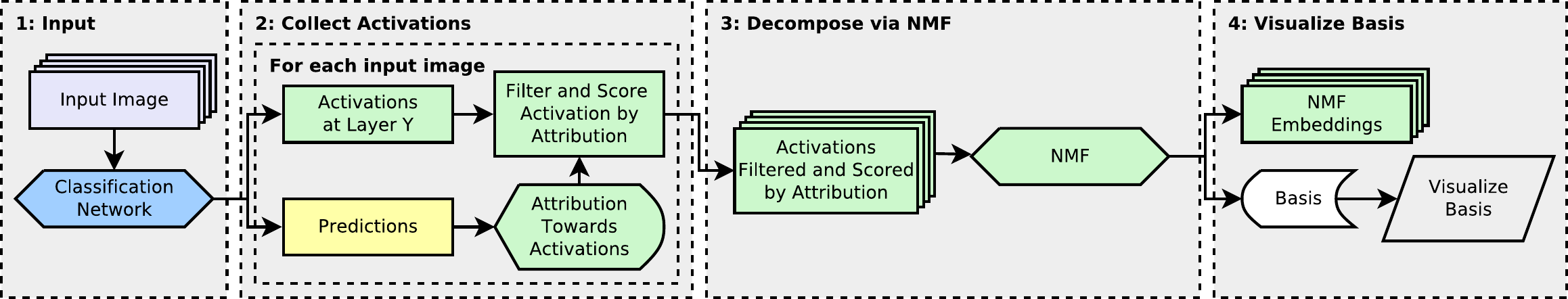}}
\caption{Method to generate explanation for a single class}
\label{fig:single_class}
\end{center}
\vskip -0.2in
\end{figure*}
Then if we extract the n (4 works good, we tested values between 1 and 10) most important concepts for a given class 'A', then combine the dataset examples for those n concepts and pass the resulting image into the model, it should predict class 'A' (i.e. we take the concepts that would be shown to the user as the explanation for class 'A' and then pass those into the model as a combined image and check whether the model predicts class 'A' for that combined image). An example of this is shown in Figure \ref{fig:basketball}.

If we explain two similar classes in isolation, then the explanations may also end up being similar and not providing enough information to understand why the model predicts one class over the other. For such a case we propose a method to contrast two classes, providing insight as to why the model predicts one class over the other.
%
%
%
%
%
%
%

%

\section{Related Work}

This work is focused on providing explanations for already trained deep neural networks, as opposed to training inherently interpretable models \cite{Nauta_2023_CVPR}.

Previous work employed attribution methods, that can be used to generate heat maps, to highlight areas in the input image that are important for the output (\cite{shrikumar2019learning}, \cite{petsiuk2018rise}). A fundamental problem with such methods is that they only highlight where in the input the model is seeing something, but do not explain what the model is seeing there. Using such attribution methods at the input layer (like its usually done), can give unreliable results. For some of those methods \cite{SanityCheckSaliencyMaps} has shown that they fail a sanity check and produce similar explanations for a trained and randomly initialized model. Those methods that passed the check, tend to give noisy explanations \cite{SmoothGrad}.

In our case, we use attribution methods (DeepLift, but we also tested using 'gradient times activation' \cite{simonyan2014deep} and SmoothGrad), but at a hidden layer, not at the input layer. That is similar to \cite{olah2018the}, where attribution methods at a hidden layer were coupled with visualization techniques, therefore answering both where the model is seeing something (attribution) and what it is seeing there (visualization). Similar to \cite{olah2018the} we also use non-negative matrix factorization (NMF) \cite{Lee1999} on the intermediate activations at a hidden layer to get the explanation down to a manageable number of concepts. But instead of using all activations, we first filter and score them by attribution, and instead of running NMF on a single image only we run it simultaneously on 500 images (we tested our method with 50 to 900 images). Using such a dataset of images to extract relevant concepts is similar to ACE and CRAFT, but we use the whole images to extract concepts instead of using image crops.
We visualize them using image crops and show them as explanation to the user.

The test we use for validating the explanation for a single class, is inspired by \cite{carter2019activation}.
There two images of two classes were combined, to change the prediction of the model to a third class (e.g. combine an image of grey whale with one of a baseball, then the model predicts great white shark for the combined image).
Finding such combinations required manual selection of a user based on insights into the model.
In our case, finding such combinations works automatically.
%
Also we do not use the whole image, but stitch crops of images.
%

%
%

\begin{table}[t]
\caption{Prediction test for the explanation. Allowing dataset crops predicted as the target class}
\label{tab:data_vis_test}
\vskip 0.15in
\begin{center}
\begin{small}
\begin{sc}
\begin{tabular}{lccccccc}
\toprule
Model & Layer & Average Pred & Predicts Target Class \\
\midrule

ResNet50 Robust  & 4.2  & 0.822 & 91.0\% \\
ResNet50    & 4.2 & 0.985 & 100.0\% \\
ResNet50 (All 1000 Classes)    & 4.2 & 0.981 & 99.6\% \\
ResNet34    & 4.2 & 0.954 & 97.0\% \\

\bottomrule
\end{tabular}
\end{sc}
\end{small}
\end{center}
\vskip -0.1in
\end{table}

\begin{table}[t]
\caption{Prediction test for the explanation. Excluding dataset crops predicted as the target class}
\label{tab:data_vis_test_exclude_target}
\vskip 0.15in
\begin{center}
\begin{small}
\begin{sc}
\begin{tabular}{lcccccc}
\toprule
Model & Layer & Average Pred & Predicts Target Class \\
\midrule

ResNet50 Robust   & 4.2 & 0.698 & 86.0\% \\
ResNet50    & 4.2 & $0.884 \pm 0.017$ & $91.4\% \pm 2.1\%$ \\
ResNet50 (All 1000 Classes)    & 4.2 & 0.892 & 91.1\% \\
ResNet34    & 4.2 & 0.850 & 86.0\% \\
\bottomrule
\end{tabular}
\end{sc}
\end{small}
\end{center}
\vskip -0.1in
\end{table}

\section{Models}

We evaluate our method on three openly available models trained for classification of natural images (ImageNet1K \cite{deng_imagenet_2009}). We use a ResNet50 and a ResNet34 \cite{resnets} from pytorch model zoo, and a ResNet50 with weights trained for adversarial robustness \cite{NEURIPS2019_6f2268bd} (in the following we refer to this model as ResNet50 Robust).
%

In our method we perform operations at a hidden layer of the model.
%
%
%
Deeper layers work better, and the penultimate layer (layer4.2) works best, as shown in Figure \ref{fig:grid_search}. And we always use the layer after relu, so that we have non-negative activations and can use NMF.

\section{Individual Class}

%
%
In this section we first describe our method and then our experimental results for explaining a single class.

\subsection{Generation Process}

We want to explain why the model predicts a specific class 'A', by extracting and visualizing the n (we usually use 4 for layer4.2, but did a grid search between 1 and 10, see Figure \ref{fig:grid_search}) main concepts that support that class. Those can be seen as the n prototypes for that class. Figure \ref{fig:single_class} shows how we extract those n concepts.
%
%
First we need a list of images.
%
We only use images where the model predicts class 'A' as majority class and we usually use 500 images (but using 100 images already performs similar, 50 is a bit worse, see Appendix \ref{sec:hyperparameters} for more information about the choice of hyperparameters).
%
%
It is important to note here that we ignore the ground truth label, we include only those and all those images where the model predicts class 'A'. That is, because excluding falsely classified images may omit crucial information necessary to understand why the model fails.

For each of the images we extract the intermediate activations at a hidden layer Y (this layer needs to be chosen manually, and that same layer is then used for the whole following process). We also investigate which layer works best, coming to the result that deeper layer work better and the penultimate layer works best (similar to the results observed in CRAFT). Then we filter and score those activations pixel wise by attribution (we compute the attribution for class 'A' towards layer Y). As attribution method we use DeepLift but also tested gradient times activation and SmoothGrad, see Appendix \ref{sec:hyperparameters}. The attribution is computed from the prediction of the model for class 'A' before softmax. As baseline for DeepLift we use a white image after normalization. The attribution method outputs something of shape (batch, channels, height, width), we then compute the mean over the channel dimension to get the contribution per pixel. Then pixels that have negative attribution are set to zero (we only include pixels that contribute positively for class 'A'), the others are multiplied by the attribution (so that pixels with higher positive attribution have higher impact).

Then in the third step, those filtered and scored activations are decomposed using NMF into n concepts. For the NMF decomposition we use scipy \cite{2020SciPy-NMeth} version 1.10.1 with the default parameters.
%
The NMF outputs a list of embeddings (we do not use those), and n vectors (the basis of the reduced space).
In the last step we visualize those n vectors (described in the next section Section \ref{sec:visualization}), and interpret those visualizations as showing n concepts the model is using to predict class 'A'.
Those visualizations would be our explanation as to why the model predicts class 'A'.

\subsection{Visualize NMF Basis Vectors}
\label{sec:visualization}

In this section we describe the method we use to visualize the n NMF basis vectors.
Each NMF basis vector, is a vector in a hidden layer of the model.
%
%
%
%
Given such a vector $v$ out of the hidden layer Y, the goal is to visualize $v$ (find an input that represents $v$).
%
%
We use dataset examples as visualization method.

As dataset examples we use use crops of a size of 74x74.
Each training image is split into 9 crops in a 3x3 grid, resulting in 11.5 million crops.
For each of those crops we compute its spatial average activation in layer Y and then the cosine similarity of the resulting vector with $v$.
Then we show those m (usually eight) crops that have the highest cosine similarity.

%
An example of this is shown in Figure \ref{fig:nmf_basis_vis_siberian_husky_from_dataset}.
Each row shows the eight most similar dataset crops for one of the NMF basis vectors (the four columns represent the four basis vectors, we used $n=4$ here).
The similarity decreases from left to right.

%
%

\begin{figure}[tb]
\vskip 0.2in
\begin{center}
\centerline{\includegraphics[width=\columnwidth]{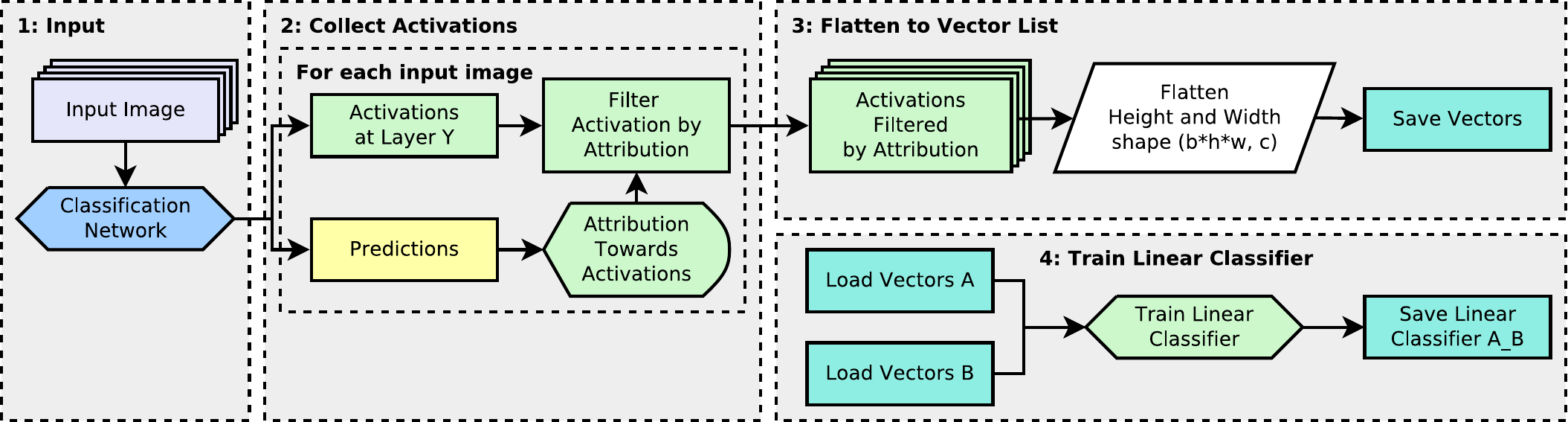}}
\caption{Contrast two classes; Data generation for the linear classifier, and training of the linear classifier}
\label{fig:contrast-vector-data}
\end{center}
\vskip -0.2in
\end{figure}

\subsection{Test Explanations}

Since we do not know what features the model is using to make its prediction (we do not have ground truths), it is difficult to validate how correct/wrong the explanation is.
%
%
For our validation test, we first combine the dataset visualization for each of the n concepts into a single image, by concatenating them in the height dimension.
Then we check how much the model predicts class 'A' for that created image and whether it predicts class 'A' as majority class.
For example, we would pass the first column of Figure \ref{fig:nmf_basis_vis_alaskan_malamute_from_dataset} (the figure shows the dataset visualizations for Alaskan Malamute) into the model and then check whether/how much the model predicts Alaskan Malamute for that image.

The result for 100 classes is shown in Table \ref{tab:data_vis_test}.
Due to computational and time constraints we only use every tenth class, resulting in 100 classes (except for the ResNet50 model, there we report the result for every tenth class as well as all 1000 classes).
%
%
Under average prediction we report the mean prediction over running the test for each of the 100 classes. Predicts target class reports how often the model predicts the target class as majority class. And we use four concepts ($n=4$).
%
%
%
The test works best for the ResNet 50 model, When using layer4.2, it works in 99.6\% of the cases and it works worst for the ResNet50 Robust model where it works in 91.0\% of the cases. In any case where this test fails, the explanation for that class should be discarded (see Section \ref{sec:case_studies} for an example).
%
%

What can be a bit unclear, is whether the model predicts the target class 'A' for the visualizations because the visualizations represent the main features the model is using to predict that class 'A', or whether it is because the model is already predicting class 'A' for the individual crops and then it does the same for the combined image.
Therefore, we also run this test while only using crops where the model is not predicting class 'A' as majority class, and those results are shown in Table \ref{tab:data_vis_test_exclude_target}.
%
%
For example, for the ResNet50 model this test works in 91.1\% of the cases when using layer4.2 (the last convolutional layer). While this test performs a bit worse compared to allowing class 'A', it makes a stronger point if it works.

\begin{figure}[tb]
    \centering
    \begin{minipage}{0.49\columnwidth}
        \centering
        \includegraphics[width=\textwidth]{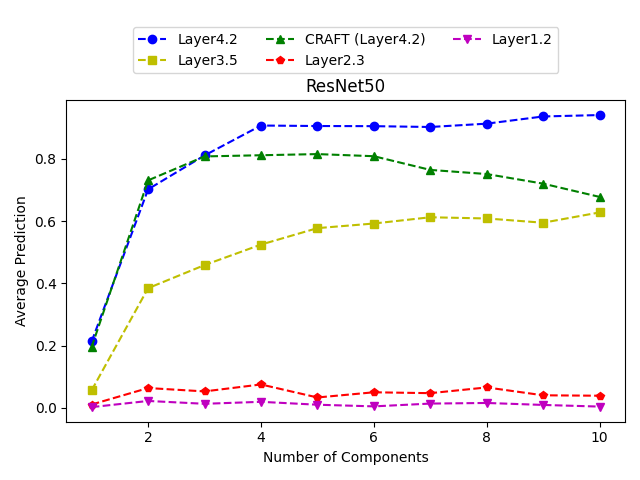}
    \end{minipage}
    \hfill
    \begin{minipage}{0.49\columnwidth}
        \centering
        \includegraphics[width=\textwidth]{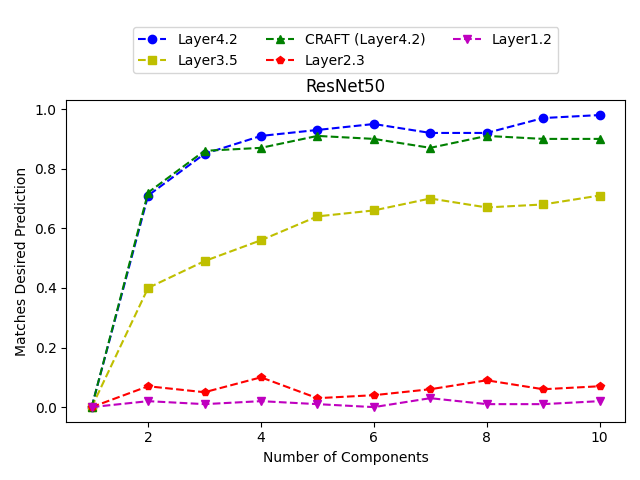}
    \end{minipage}
    \caption{Testing different number of components in the NMF and different hidden layers. The target class is excluded in sampling of the concepts}
    \label{fig:grid_search}
\end{figure}

In Figure \ref{fig:grid_search} we test what effect of the number of components in the NMF and the chosen hidden layer has. The left plot shows the average prediction for the target class and the right plot shows how often the explanation gets predicted as the target class (would correspond to 'predicts target class' in Table \ref{tab:data_vis_test_exclude_target}). In both cases higher is better.
We also compare to CRAFT. We use CRAFT with the configuration provided in their code example \cite{CRAFT_Code} (penultimate layer, 10 components in the NMF) and use 500 images as with our method (the code example of CRAFT used 300 images). For our method, we tested four layers (layer1.2, layer2.3, layer3.5 and layer4.2) at different depths of the model. We can see that deeper layer work better, layer1.2 and layer2.3 do not work and the penultimate layer (layer4.2) works best.

When we change the number of components, we need to rerun our method. With CRAFT we only run it once with 10 components, and then use their method to score the importance of each component with relation to the target class to extract the n most important components.
%
Our method slightly outperforms CRAFT when using four or more components and slightly outperforms it overall. The average prediction for CRAFT goes down when using more than 6 components. That might be because the components in CRAFT represent all the activations that are present in the images, and not only those that are predictive for the target class. Therefore, using too many of the components might add features that are not predictive for the class and therefore reduce the prediction.

In the right plot, when using a single component we never match the desired target class (get zero score) by construction, because for visualizing the concepts we exclude those dataset crops where the model predicts the desired target class as majority class. Therefore, when using a single component, we only have a single dataset crop which is never predicted as the target class.

For all other experiments we always use four components (four concepts) for layer4.2 because using more components does not improve the prediction much, and therefore we choose to have fewer concepts.

\section{Class Contrasting}
\label{sec:class_contrasting}

In this section we look at contrasting classes, i.e. investigating why the model predicts class 'A' over 'B'.
A problem that can arise from only explaining each class individually, is that the explanation for two classes can be very similar (e.g. Alaskan Malamute and Siberian Husky, see Section \ref{sec:case_studies}), where the explanations look virtually indistinguishable.
In that example, the explanation for Alaskan Malamute actually fails (for the dataset visualization the model predicts an average of 0.26 for Alaskan Malamute, and the majority class is Siberian Husky with a prediction of 0.42).
Which means we only get an explanation for Siberian Husky, but none for Alaskan Malamute.

Therefore in the next step we do not explain Alaskan Malamute in isolation, but contrast it with the class it failed with (Siberian Husky). That results in a much better explanation for Alaskan Malamute, in the sense that the model now predicts it as majority class with a confidence of 0.81.

In the following we describe our method and results for such contrasting between two classes.

%
%

%


\begin{table}[t]
\caption{Post-softmax predictions when shifting activations in the latent space}
\label{tab:shifting}
\vskip 0.15in
\begin{center}
\begin{small}
\begin{sc}
\begin{tabular}{lcccccc}
\toprule
Model & Layer & Default Pred & Shifted Pred \\
\midrule
ResNet50 & 4.2    & 0.000 $\pm$ 0.000 ($\pm$0.002) & 0.995 $\pm$ 0.000 ($\pm$0.036) \\
ResNet50 & 3.5    & 0.000 $\pm$ 0.000 ($\pm$0.002) & 0.375 $\pm$ 0.004 ($\pm$0.420) \\

\bottomrule
\end{tabular}
\end{sc}
\end{small}
\end{center}
\vskip -0.1in
\end{table}

\subsection{Method}

The idea is to train a linear classifier (single linear layer with a scalar bias and a sigmoid in the end) for binary classification between activations that are predictive for class 'A' and those that predictive for class 'B'.
%
%
Then the normal of the hyperplane spanned by the linear classifier would ideally point from class 'A' to 'B', and those activations that are far along the hyperplane normal in the direction of class 'A' would be predictive for class 'A' vs 'B'.
Its a similar approach as before for explaining a single class, the only difference is that instead of using attribution to filter and score the activations, we use the linear classifier to filter and score the activations. The rest is the exact same.
Figure \ref{fig:contrast-vector-data} shows our approach to generate the hyperplane separating class 'A' and 'B'.
This hyperplane is used in the next step to extract the n most important concepts separating class 'A' from 'B'.
%
%

\subsubsection{Hyperplane}

First we describe Figure \ref{fig:contrast-vector-data}, which shows the generation process of the linear classifier.
Step 1 to 3 are done individually per class, step 4 is run for all combinations of any two classes. In the case of ImageNet with 1000 classes we would get roughly one million linear classifiers (we exclude the combination of a class with itself).
Step 1 and 2 are similar to Figure \ref{fig:single_class}, the only difference is that we do not scale the activations by the attribution, but filter only.
And the filtering is a bit more aggressive. Instead of filtering out only those pixels that have negative attribution, we also filter out all those pixels that have less than 0.25 of the maximum attribution over all pixels of that image.
The reason why we choose a higher cutoff, is that previously the activations would be scaled by the attribution, so that those with lower attribution would have lower impact on the final concepts.
But now all the activations that pass the filtering step have the same impact on the linear classifier, therefore we use a higher cutoff.

In the next step the activations of the pixels remaining after the filtering are saved (i.e. we save a list of vectors per class, each vector representing a pixel in the activations at layer Y).
%
%
Then in the last step we train a linear classifier to do binary classification between two such lists of vectors (one list for class 'A' and one for class 'B').
The linear classifier is always trained for 100 epochs, running the whole dataset at once (no mini-batching).
As optimizer we use stochastic gradient descent \cite{10.1214/aoms/1177729586}, \cite{10.1007/978-3-7908-2604-3_16} with a learning rate of 0.01.

\begin{figure}[tb]
    \centering
    \begin{minipage}{0.49\columnwidth}
        \centering
        \includegraphics[width=\textwidth]{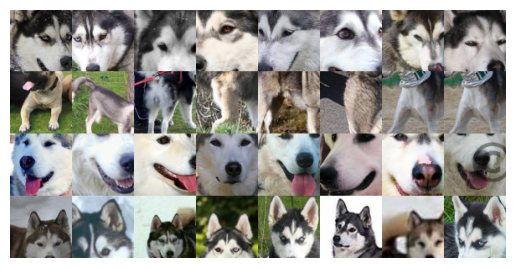}
        \caption{Visualization for the 4 NMF components for Alaskan Malamute}
        \label{fig:nmf_basis_vis_alaskan_malamute_from_dataset}
    \end{minipage}
    \hfill
    \begin{minipage}{0.49\columnwidth}
        \centering
        \includegraphics[width=\textwidth]{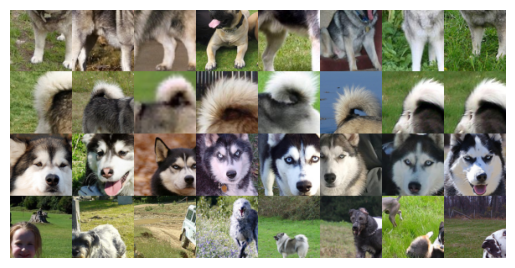}
        \caption{Visualization for the 4 NMF components pro Alaskan Malamute vs Siberian Husky}
        \label{fig:nmf_basis_vis_alaskan_malamute_vs_husky_from_dataset}
    \end{minipage}
    
\end{figure}



\begin{figure}[tb]
    \centering
    \begin{minipage}{0.49\columnwidth}
        \centering
        \includegraphics[width=\textwidth]{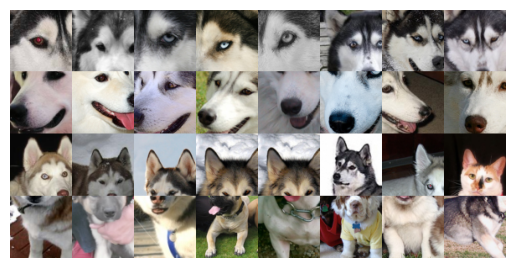}
        \caption{Visualization for the 4 NMF components for Siberian Husky}
        \label{fig:nmf_basis_vis_siberian_husky_from_dataset}
    \end{minipage}
    \hfill
    \begin{minipage}{0.49\columnwidth}
        \centering
        \includegraphics[width=\textwidth]{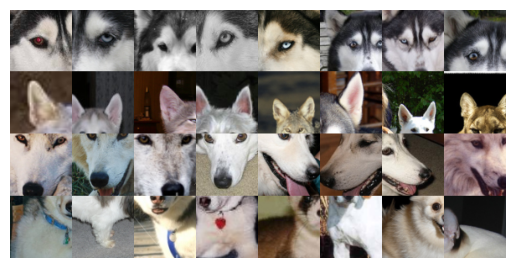}
        \caption{Visualization for the 4 NMF components pro Siberian Husky vs Alaskan Malamute}
        \label{fig:nmf_basis_vis_siberian_husky_vs_malamute_from_dataset}
    \end{minipage}
    \label{fig:nmf_basis_vis_husky_from_dataset}
\end{figure}


\subsubsection{Extract Separating Concepts}

Now we describe how we extract the n concepts for the contrasting.
%
%
The method is similar to the method explaining a single class, shown in Figure \ref{fig:single_class}.
The only difference is in step 2, where we now scale and filter the activations by the linear classifier, instead of using attribution.

We run the linear classifier (without the sigmoid, so only the dot product between the weight vector and the pixel and adding the scalar bias) pixel wise on all the intermediate activations we get at layer Y.
%
%
Let $w$ be the weight vector, $b$ the scalar bias and $x$ the pixel of the intermediate activations at layer Y, then we do: $z = w \cdot x + b$, where the output $z$ is a scalar value.
Then for each pixel we do: $x' = \text{max}(z, 0) \cdot x$, that is we scale the activations by $z$ if $z$ is larger than zero, otherwise we set those activations to zero.
We only keep those activations that are on the side of the hyperplane that is predictive for class 'A', and scale those remaining activations by how far along the hyperplane normal they are.
The remaining two steps are the exact same as for explaining a single class.
The modified activations are decomposed via NMF into n concepts, and those n concepts are then visualized.

\subsection{Results}

\subsubsection{Shifting Test}
\label{sec:shifting_test}

%
%
The test we propose for the contrasting, tests the direction of the hyperplane. Ideally, the hyperplane normal should point from class 'B' towards class 'A'. This means that if we would start from an image of class 'B', and then translate the activations in the hidden layer along the hyperplane normal, we would like the model to swap prediction to class 'A'. We do the translation using ten offset values (those are fixed and the same for all classes and layers) and report the result of the offset value where the prediction for class 'A' is highest. We do not simply use a single large value, because the prediction for class 'A' may decrease again when going too far along the hyperplane normal (this happens for layer3.5).

The results from this translation test are shown in Table \ref{tab:shifting}. Default pred as well as shifted pred both show the results for class 'A'. Default pred shows the mean prediction for class 'A' for the original activations (when not adjusting anything). Shifted pred shows the mean prediction for class 'A' when translating the activations along the hyperplane normal.
%
%
%
We have too many classes to run all combinations time wise, instead we use each of the 1000 classes as starting class ('B'), and for each of them we choose only 10 random classes ('A') to shift towards. Then we repeat this 5 times and report mean $\pm$ standard deviation over the 5 runs as well as ($\pm$ standard deviation over all samples).

\begin{table*}[t]
\caption{Prediction for the original and stitched images. Both when inserting the grass patch, as well as inserting zeros instead of that grass patch}
\label{tab:example_preds_malamute_husky}
\vskip 0.15in
\begin{center}
\begin{small}
\begin{sc}
\begin{tabular}{lcccccc}
\toprule
Class & Original Pred & Grass Inserted & Black Inserted  \\
\midrule
Siberian Husky    & 0.165 & 0.117 & 0.143  \\
Alaskan Malamute   & 0.080 & 0.100 & 0.071 \\
\toprule
  & concept 0 & concept 1 & concept 2 & concept 3 \\
\midrule
Siberian Husky     & 0.176 & 0.162  & 0.444 & 0.117 \\
Alaskan Malamute    & 0.127 & 0.155  & 0.325 & 0.080 \\
\bottomrule
\end{tabular}
\end{sc}
\end{small}
\end{center}
\vskip -0.1in
\end{table*}

\section{Case Studies}
\label{sec:case_studies}

\subsection{Alaskan Malamute vs Siberian Husky}

In this section we investigate why the ResNet50 Robust model classifies an image as Alaskan Malamute vs Siberian Husky and vice versa (in the appendix we show more examples).
We apply the method at layer4.2.
Both classes are dog breeds that look very similar.
%
%
Figure \ref{fig:nmf_basis_vis_alaskan_malamute_from_dataset} shows visualizations from dataset crops for the four most important concepts supporting Alaskan Malamute, whereas Figure \ref{fig:nmf_basis_vis_siberian_husky_from_dataset} shows the same for Siberian Husky.
We can see that the model mostly uses the dog face to predict either class.
%
%
The explanation for Alaskan Malamute is a failure case, since the model predicts Siberian Husky as majority class for the four concepts (0.26 for Alaskan Malamute and 0.42 for Siberian Husky).
Both sets of images (Alaskan Malamute and Siberian Husky) look very similar, just based on this it is not possible to understand why the model predicts one class versus the other.

Next we can contrast both classes (using our method described in Section \ref{sec:class_contrasting}), and look at the four most important concepts separating one class from the other.
Now we can see differences between the visualizations of both classes, and the contrasted explanation (Figure \ref{fig:nmf_basis_vis_alaskan_malamute_vs_husky_from_dataset}) gets predicted as Alaskan Malamute with a confidence of 0.81 (much higher compared to the 0.26 when explaining Alaskan Malamute in isolation). For Alaskan Malamute, there is now only one component showing the dog head, as shown in the third row of Figure \ref{fig:nmf_basis_vis_alaskan_malamute_vs_husky_from_dataset}. Interestingly, three out of four components show grass, which potentially points to a bias in the model, since it should not matter whether the dog is standing on grass or e.g. snow.

%
%
%
%
%

\begin{figure}[tb]
    \centering
    \begin{minipage}{0.25\columnwidth}
        \centering
        \includegraphics[width=\textwidth]{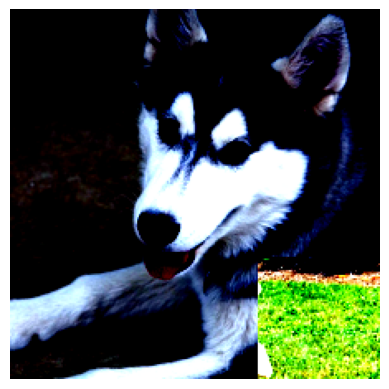}
    \end{minipage}
    \hfill
    \begin{minipage}{0.25\columnwidth}
        \centering
        \includegraphics[width=\textwidth]{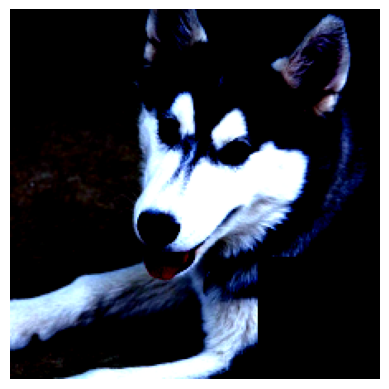}
    \end{minipage}
    \caption{Example of inserting the grass patch (or zeros) into an originally as husky predicted image (image shown after normalization)}
    \label{fig:husky_person_inserted}
\end{figure}

To test this bias, we insert an image of grass into the first 128 training images predicted as Siberian Husky (always at the bottom right), and then check whether the model changes its prediction to Alaskan Malamute.
%
%
One of those 128 modified images is shown in Figure \ref{fig:husky_person_inserted}, once with the grass image inserted and once with zeros inserted. The image is shown after normalization.
Table \ref{tab:example_preds_malamute_husky} shows the average prediction for Siberian Husky and Alaskan Malamute for those 128 images. Originally the model predicts more Siberian Husky (0.165). After inserting the grass image into the lower right of each of the images, the model predicts less Husky (0.117) and more Malamute (0.100 vs originally 0.080). Since we occlude part of the image by inserting the grass image, we also test the prediction if we insert zeros (black) instead. In that case the prediction does not change much, and slightly decreases for both Husky as well as Malamute.

The bottom half of the table reports the prediction when inserting a concept from contrasting Malamute with Husky (from Figure \ref{fig:nmf_basis_vis_alaskan_malamute_vs_husky_from_dataset}). E.g. concept 0 would report the prediction when inserting the first image of the first row from Figure \ref{fig:nmf_basis_vis_alaskan_malamute_vs_husky_from_dataset}, concept 1 would be the first image from the second row. concept 1 (the tail image) works best, inserting it slightly decreases the prediction for Husky and raises the prediction for Malamute close to that of Husky.

\section{Limitations}
\label{sec:limitations}

%
%
%
%
%
%
We only tested the method on convolutional neural networks. Further, for a class where the test fails, the method should not be trusted (as shown in the case study where it fails for Malamute). That test fails in 9\% of the classes for the worst model. The stricter test (excluding dataset crops predicted as the target class) fails in 9\% of the classes for the best model, and for the worst model it fails in 14\% of the classes. That stricter test also requires a dataset that contains a variety of concepts from different classes. Because it will likely not work for e.g. binary classification between two very dissimilar classes (like airplane vs dog). Because its likely not possible to visualize concepts that explain the prediction for dog from airplane images.


Also, visualizing the concepts via dataset examples (as opposed to using iterative gradient descent to synthesize them) may result in concepts that are misleading. For example, if a concept shows a dog head, does it mean that concept represents the whole head, or maybe only the nose or eyes? We discuss that more in Appendix \ref{sec:generate_visualization_grad_desc}, \ref{sec:bcc_vs_trichoblastoma}.

%
%
%


%
%

\section{Impact Statement}
\label{sec:impact_statement}

This paper presents work whose goal is to advance the field of Machine Learning, especially in the field of explainable AI. Obtaining a better understanding of the behavior of the models could have a positive effect.
%

\section{Acknowledgments}

R.H. is funded by the Deutsche Forschungsgemeinschaft (DFG, German Research Foundation) - Projectnumber 459360854 (DFG FOR 5347 Lifespan AI). D.O.B. acknowledges the financial support by the Federal Ministry of Education and Research (BMBF) within the T!Raum-Initiative "\#MOIN! Modellregion Industriemathematik" in the sub-project "MUKIDerm".

\bibliographystyle{plainnat}
\bibliography{example_paper}

\newpage
\appendix
\FloatBarrier

\FloatBarrier

\section{Hardware and Software}
\label{sec:hardware_and_software}

On the software side we use pytorch \cite{paszke2019} version 1.13.1 with cuda version 11.6 and torchvision \cite{torchvision2016} version 0.14.1.
For the attribution methods (for DeepLift, SmoothGrad and gradient times activation) we use the captum library \cite{kokhlikyan2020captum} version 0.6.0.
For NMF we use scipy \cite{2020SciPy-NMeth} version 1.10.1.

For the hardware, we ran the experiments on a Linux server with 8 Nvidia RTX 2080Ti GPUs and we also used another Linux server with 4 Nvidia RTX A6000 GPUs. But the experiments are all doable on a single RTX 2080Ti GPU.

\section{Computational Requirements}
\label{sec:computational_requirements}

In the following, all times are reported running on a single RTX 2080Ti GPU with an AMD EPYC 7262 8-Core Processor (32 CPUs).
Extracting the NMF components for all 1000 classes for a single model and layer takes 15 hours for the ResNet50 model. Visualizing the concepts for all 1000 classes adds another 14 hours (around 50 seconds per class), here the performance cost comes from comparing each of the n NMF components with each of the 11.5 million dataset crops (vector wise cosine similarity).

For the class contrasting, generating the activations for all 1000 classes takes 3 hours (step 1 to 3 shown in Figure \ref{fig:contrast-vector-data}). Training the pairwise linear classifiers takes 35 hours. Using those trained classifiers to contrast two classes then takes around 90 seconds (40 seconds for generating the NMF components and another 50 seconds for the visualizing the concepts).

\section{concept Intruder Detection}

\begin{table}[t]
\caption{Detecting an intruder concept}
\label{tab:appendix_intruder_detection}
\vskip 0.15in
\begin{center}
\begin{small}
\begin{sc}
\begin{tabular}{cccccccc}
\toprule
Number of Components & Detection Accuracy \\
\midrule

4 & 77\% \\
8 & 76\% \\

\bottomrule
\end{tabular}
\end{sc}
\end{small}
\end{center}
\vskip -0.1in
\end{table}

\begin{figure}[tb]
\vskip 0.2in
\begin{center}
\centerline{\includegraphics[width=\columnwidth]{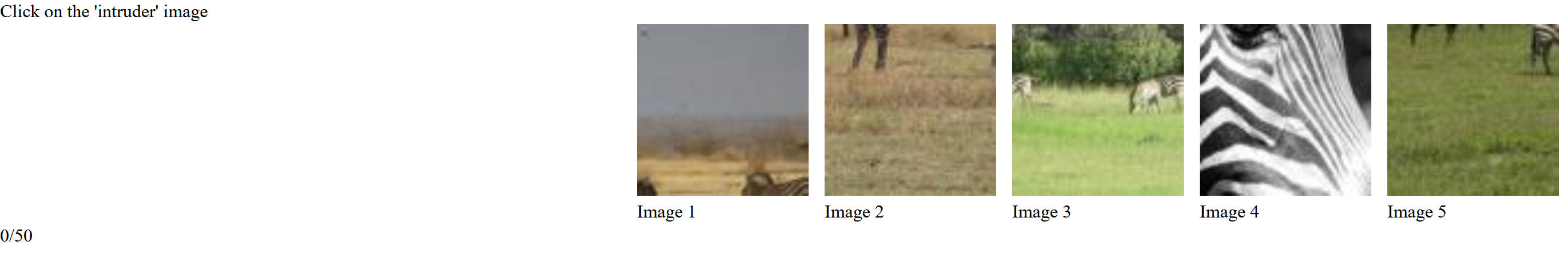}}
\caption{Intruder detection experiment}
\label{fig:intruder_detection}
\end{center}
\vskip -0.2in
\end{figure}

In order of testing the human interpretability of the extracted concepts, we asked 4 persons affiliated with the authors of this paper (co-workers) to detect an intruder image out of five images. They are shown five images, from which four represent one concept and one represents another concept but from the same class. The goal is to identify that one image that represents the other concept. The idea is to check how human interpretable the concepts are, i.e. if they all look the same, then their usefulness as explanation will be low. This is done for fifty intruder images (fifty steps), each time selecting a new random class out of the 1000 classes and two random concepts (one for the four images and another concept for the intruder image). And this experiment is done using the ResNet50 model and layer4.2.

The result is shown in Table \ref{tab:appendix_intruder_detection}. Its run twice, one time using an explanation with four concepts and one time using an explanation with eight concepts. This was to check whether the explanation becomes less interpretable if more concepts are used (eight vs four). Because there may be more overlap between eight concepts than there is with four. But the result is similar (77\% detection accuracy for four concepts and 76\% for eight concepts). Random guessing would be around 20\%, therefore we see that as a confirmation that our method provides human interpretable explanations.

\section{Hyperparameters}
\label{sec:hyperparameters}

\subsection{Number of Samples}

\begin{figure}[tb]
    \centering
    \begin{minipage}{0.49\columnwidth}
        \centering
        \includegraphics[width=\textwidth]{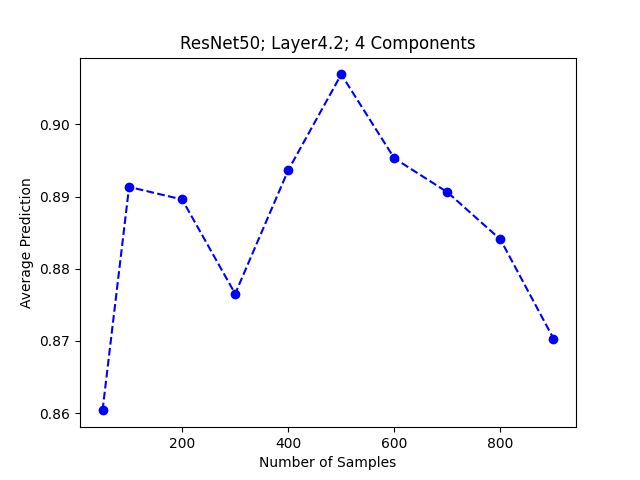}
    \end{minipage}
    \hfill
    \begin{minipage}{0.49\columnwidth}
        \centering
        \includegraphics[width=\textwidth]{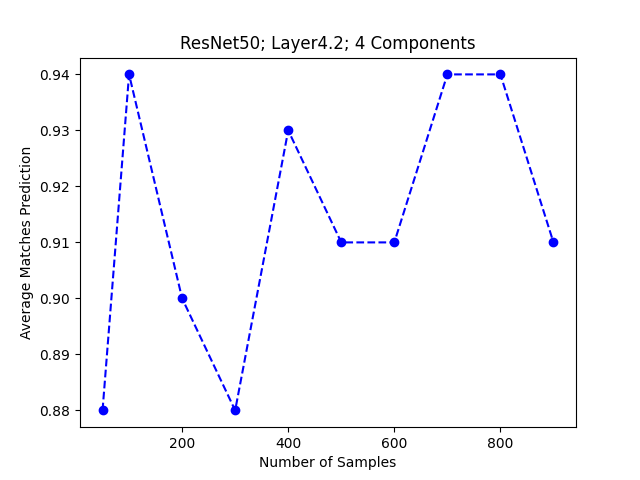}
    \end{minipage}
    \caption{Testing different number of components in the NMF and different hidden layers}
    \label{fig:grid_search_number_of_samples}
\end{figure}

In Figure \ref{fig:grid_search_number_of_samples} we plot how our method performs depending on the number of samples that is used to generate the explanation. We vary the number of samples between 50, 100 and then in steps of 100 till 900. We keep the other hyperparameters (layer4.2, 4 components, DeepLift as attribution method) fixed. The left plot shows the average prediction for the class that should be explained and the right plot shows how often the explanation is predicted as that target class as majority class. And we exclude patches that are predicted as the target class in sampling the closest dataset patches for the explanation.

The standard deviation in the prediction for 500 samples is 0.02 (as shown in Table \ref{tab:data_vis_test_exclude_target}). Assuming its the same for the other data points as well, then the results for 100 to 900 samples are roughly the same, with the result for 50 samples being a bit lower. Therefore, we would recommend to have at least 100 samples available per class.

\subsection{Attribution Method}

\begin{table}[t]
\caption{Prediction test for the explanation. Allowing dataset crops predicted as the target class}
\label{tab:appendix_attribution_methods}
\vskip 0.15in
\begin{center}
\begin{small}
\begin{sc}
\begin{tabular}{lclccccc}
\toprule
Model & Layer & Attribution Method & Samples & Average Pred & Predicts Target Class \\
\midrule

ResNet50   & 4.2 & DeepLift & 50 & \textbf{0.976} & 98.0\% \\
ResNet50   & 4.2 & DeepLift plus SmoothGrad& 50 & 0.972 & \textbf{99.0\%} \\
ResNet50   & 4.2 & Gradient times Activation& 50  & 0.975 & 98.0\% \\

\hline

ResNet50   & 3.5 & DeepLift & 500 & \textbf{0.524} & 56.0\% \\
ResNet50   & 3.5 & Gradient times Activation& 500 & 0.497 & \textbf{57.0\%} \\

\bottomrule
\end{tabular}
\end{sc}
\end{small}
\end{center}
\vskip -0.1in
\end{table}

\begin{table}[t]
\caption{Prediction test for the explanation. Excluding dataset crops predicted as the target class}
\label{tab:appendix_attribution_methods_exclude_target}
\vskip 0.15in
\begin{center}
\begin{small}
\begin{sc}
\begin{tabular}{lclccccc}
\toprule
Model & Layer & Attribution Method & Samples & Average Pred & Predicts Target Class \\
\midrule

ResNet50   & 4.2 & DeepLift & 50  & \textbf{0.860} & \textbf{88.0\%} \\
ResNet50   & 4.2 & DeepLift plus SmoothGrad & 50  & 0.848 & 87.0\% \\
ResNet50   & 4.2 & Gradient times Activation & 50  & \textbf{0.860} & \textbf{88.0\%} \\

\hline

ResNet50   & 3.5 & DeepLift & 500 & \textbf{0.837} & \textbf{85.0\%} \\
ResNet50   & 3.5 & Gradient times Activation& 500 & 0.799 & \textbf{85.0\%} \\

\bottomrule
\end{tabular}
\end{sc}
\end{small}
\end{center}
\vskip -0.1in
\end{table}

In Table \ref{tab:appendix_attribution_methods} and Table \ref{tab:appendix_attribution_methods_exclude_target} we investigate the effect of using different attribution methods. This experiment is run using the ResNet50 model for layer4.2 and layer3.5. As attribution methods we use DeepLift, DeepLift plus SmoothGrad (using 20 samples and a standard deviation in the gaussian noise of 0.25) and gradient times activation. In layer4.2 all three methods perform similar, with DeepLift plus SmoothGrad performing slightly worse. In layer3.5 we only compare DeepLift and gradient times activation, here we want to check whether the non-linearity becomes a problem when using gradient times activation. But gradient times activation only performs slightly worse compared to using DeepLift (slightly lower prediction). Overall, simply using gradient times activation can also be used, especially in the penultimate layer (layer4.2), where there is no non-linearity anyway.

\subsection{Shifting Test Prediction over Offset}

\begin{figure}[tb]
    \centering
    \begin{minipage}{0.49\columnwidth}
        \centering
        \includegraphics[width=\textwidth]{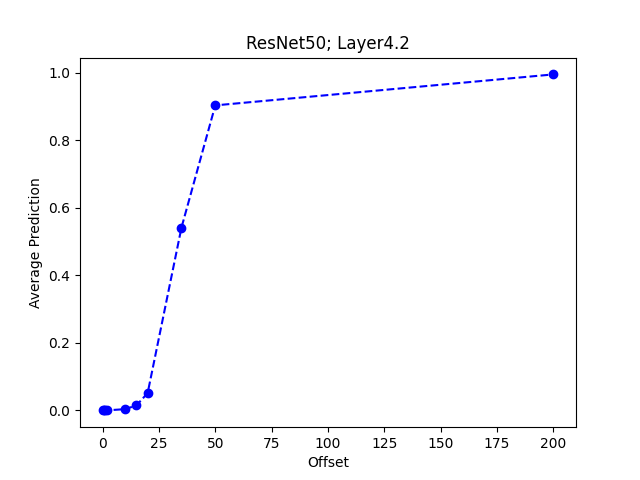}
    \end{minipage}
    \hfill
    \begin{minipage}{0.49\columnwidth}
        \centering
        \includegraphics[width=\textwidth]{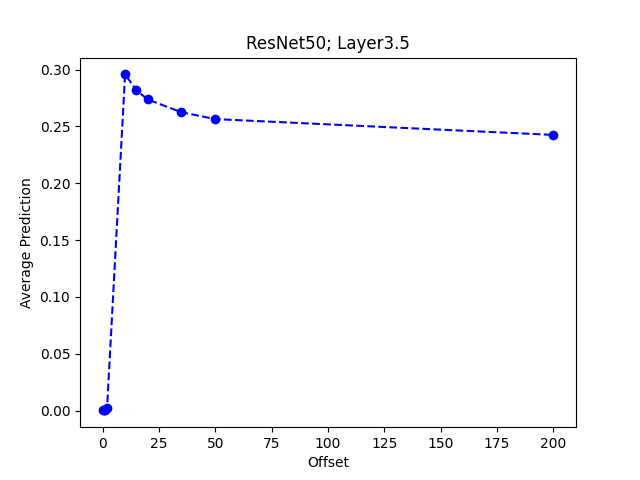}
    \end{minipage}
    \caption{Plotting the average prediction for class 'A' (the target class). The x-axis shows how far the activations are shifted along the hyperplane normal pointing from class 'B' towards 'A'}
    \label{fig:shifting_test}
\end{figure}

Figure \ref{fig:shifting_test} illustrates the offsets used in the shifting test in Table \ref{tab:shifting} in the main paper. For layer4.2, we could simply use a very large offset value, but for layer3.5 the prediction decreases again when the offset value is becomes too large.

\subsection{Number of Components and Hidden Layer}

\begin{figure}[tb]
    \centering
    \begin{minipage}{0.49\columnwidth}
        \centering
        \includegraphics[width=\textwidth]{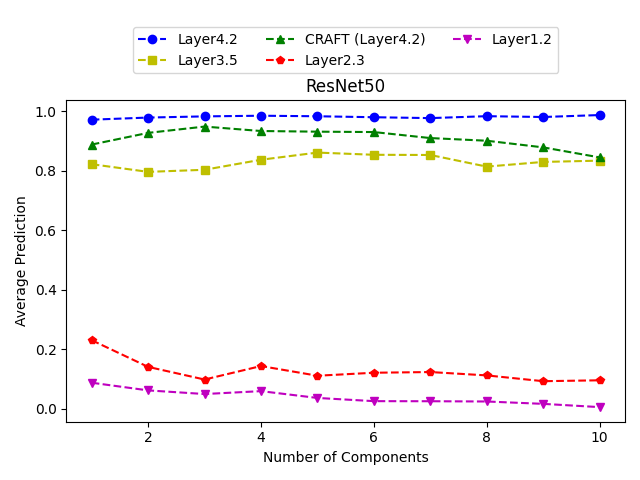}
    \end{minipage}
    \hfill
    \begin{minipage}{0.49\columnwidth}
        \centering
        \includegraphics[width=\textwidth]{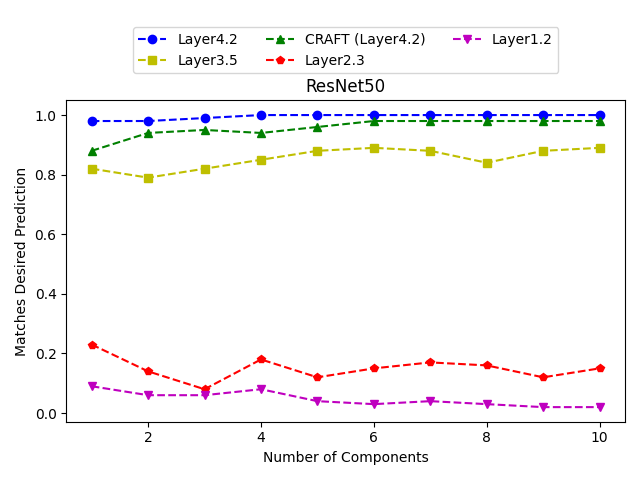}
    \end{minipage}
    \caption{Testing different number of components in the NMF and different hidden layers. The target class is allowed in sampling of the concepts}
    \label{fig:grid_search_allow_target_class}
\end{figure}

Figure \ref{fig:grid_search_allow_target_class} shows the same results as \ref{fig:grid_search} in the main paper, but allows concepts that are predicted as the target class. The result is similar, the deeper the layer the better and our method slightly outperforms CRAFT. And again layer1.2 and layer2.3 do not work (the prediction is very low).

\section{Generate Visualizations via Gradient Descent}
\label{sec:generate_visualization_grad_desc}

For this section and the next section \ref{sec:bcc_vs_trichoblastoma} we use an in-house model trained to segment tumor between 37 classes in digital pathology tissue images. It uses a modified UNet with a ResNet34 backbone and in the following we use layer3.2 (that would be 21 layers into the ResNet34). The dataset images have a size of 1536x1536 and the concepts have a size of 512x512. We use 60 samples (instead of 500) and as attribution method we use DeepLift plus SmoothGrad. And we use $n=6$ components in the NMF (i.e. we use six concepts in the explanation).

\begin{figure}[tb]
\vskip 0.2in
\begin{center}
\centerline{\includegraphics[width=\columnwidth]{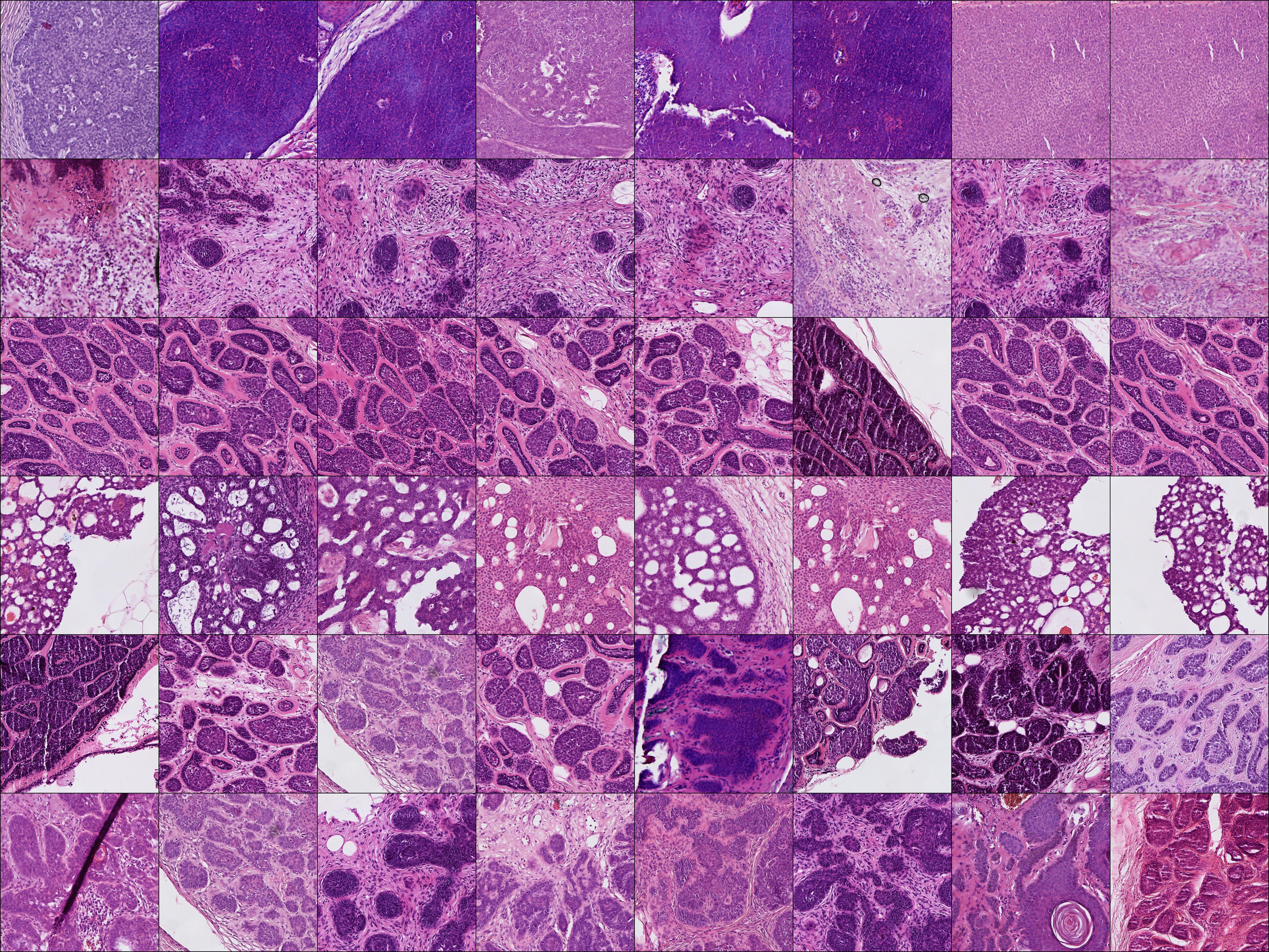}}
\caption{Dataset visualization for the 6 NMF components for BCC (best viewed zoomed in)}
\label{fig:nmf_basis_vis_from_data}
\end{center}
\vskip -0.2in
\end{figure}

In the main paper we only showed examples where we generate the visualizations by sampling crops from validation images. We can also synthesize them via gradient descent though, similar to \cite{olah2017feature} or \cite{nguyen2016synthesizingpreferredinputsneurons}. For the following digital pathology example, we also generate the visualizations for each of the six concepts via gradient descent in the latent space of a GAN (generative adversarial network). Such visualizations might be more faithful compared to sampling dataset examples. For example in Figure \ref{fig:nmf_basis_vis_3} the last third and the last two images show palisading cell edges with different orientations (the first one shows an upper right edge, the next one a lower right edge and the last one a left edge). But in the corresponding dataset visualizations shown in Figure \ref{fig:nmf_basis_vis_from_data} (third row and last two rows), there is no edge orientation visible, they simply show BCC islands. Meaning, from the dataset visualizations alone, we could not see that the model is actually looking at the edge of the BCC island that has a specific orientation (from only the dataset example we could think that its looking for the islands themselves, or for a right edge instead of a left edge).

\section{Basal Cell Carcinoma vs Trichoblastoma}

\label{sec:bcc_vs_trichoblastoma}


\begin{figure*}[tb]
\vskip 0.2in
\begin{center}
\centerline{\includegraphics[width=\columnwidth]{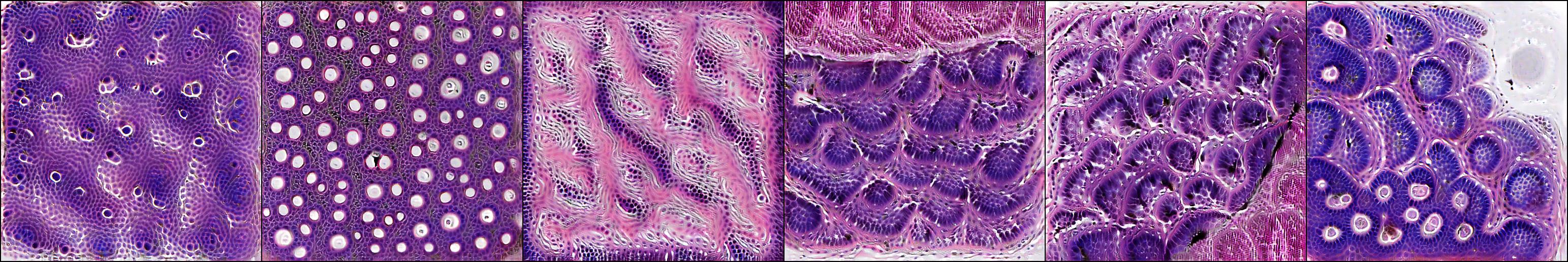}}
\caption{Visualization for the 6 NMF components for Trichoblastoma}
\label{fig:nmf_basis_vis_3_trichoblastoma}
\end{center}
\vskip -0.2in
\end{figure*}

\begin{figure*}[tb]
\vskip 0.2in
\begin{center}
\centerline{\includegraphics[width=\columnwidth]{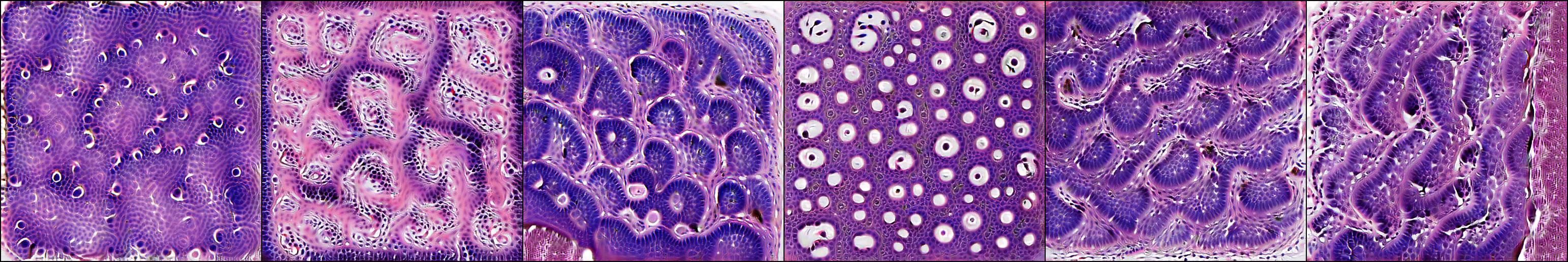}}
\caption{Visualization for the 6 NMF components for BCC}
\label{fig:nmf_basis_vis_3}
\end{center}
\vskip -0.2in
\end{figure*}

\begin{figure*}[tb]
\vskip 0.2in
\begin{center}
\centerline{\includegraphics[width=\columnwidth]{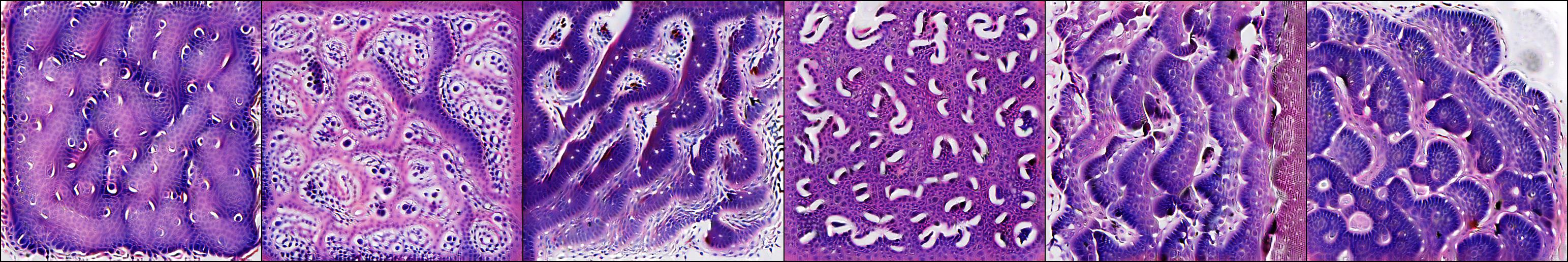}}
\caption{Visualization for the 6 NMF components pro BCC vs Trichoblastoma}
\label{fig:nmf_basis_vis_3_bcc_vs_trichoblastoma}
\end{center}
\vskip -0.2in
\end{figure*}

\begin{figure}[tb]
\vskip 0.2in
\begin{center}
\centerline{\includegraphics[width=\columnwidth]{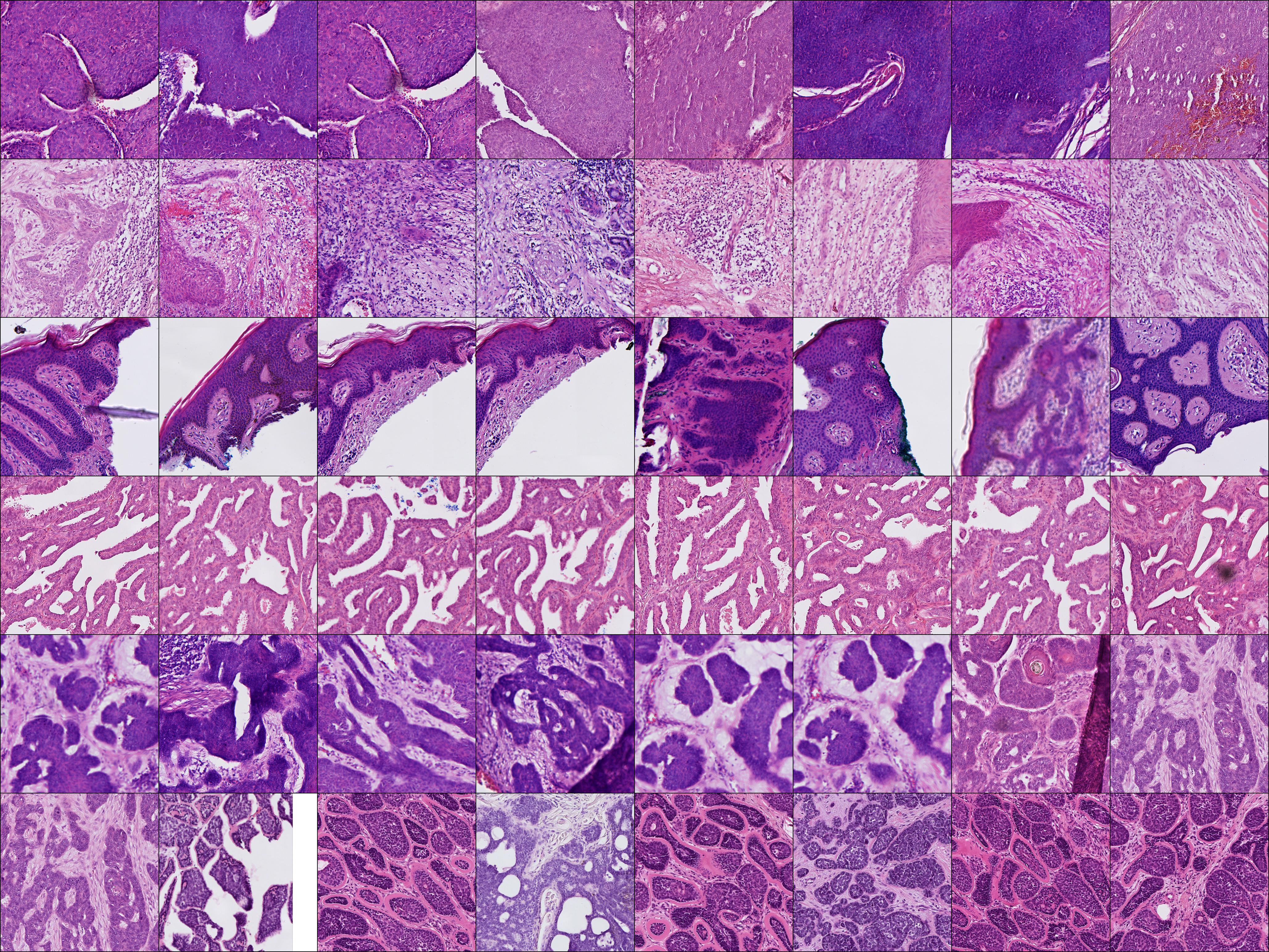}}
\caption{Dataset visualization for the 6 NMF components for BCC vs Trichoblastoma (best viewed zoomed in)}
\label{fig:1_vs_32_nmf_basis_vis_from_data}
\end{center}
\vskip -0.2in
\end{figure}

Basal Cell Carcinoma (BCC) as well as Trichoblastoma look clinically very similar to each other \cite{Patel2020}. Human experts as well as the AI model can separate them, and in this section we investigate why the model predicts Basal Cell Carcinoma over Trichoblastoma. First we can look at the main concepts the model uses to predict BCC shown in Figure \ref{fig:nmf_basis_vis_3} and compare them to the main concepts the model uses to predict Trichoblastoma shown in Figure \ref{fig:nmf_basis_vis_3_trichoblastoma}. The problem is that the concepts look very similar (which makes sense since the two diseases look clinically similar), there is no single concept that is present for one class but not the other. For example the first concept from the left for Trichoblastoma looks similar compared to the first from the left for BCC. The second one for Trichoblastoma looks similar to the fourth for BCC and the third looks similar to the second. And the last three concepts for Trichoblastoma show palisading cell edges with different orientations, similar to the third and last two concepts for BCC. So those visualizations are not enough to spot a difference between Trichoblastoma and BCC.

In the next step we look at the visualizations pro BCC we get from contrasting the two classes. Figure \ref{fig:nmf_basis_vis_3_bcc_vs_trichoblastoma} shows the 6 main concepts that are pro BCC vs Trichoblastoma. The second, third and fourth concept look differently from the 6 main concepts pro Trichoblastoma (so here we can now see a difference).
%
%
For interpreting those 3 concepts we can look at dataset examples, shown in Figure \ref{fig:1_vs_32_nmf_basis_vis_from_data}. The fourth concept shows white areas inside tumor, which could resemble cleavage (in BCC the tumor is often separated from the surrounding tissue by white area). The third concept shows epidermis, which is also interesting since according to \cite{Patel2020} BCC is sometimes attached to the epidermis, whereas Trichoblastoma never is.


For the shifting test, shifting the activations in the latent space in the direction of the hyperplane normal works both ways. Here we reach an average maximum prediction for BCC of 0.721 (at that point the prediction for Trichoblastoma is 0.481). And the other way around, we reach an average maximum prediction for Trichoblastoma of 0.673 and at that point the prediction for BCC is 0.017.

\section{Examples ResNet50}

\begin{figure}[tb]
\vskip 0.2in
\begin{center}
\centerline{\includegraphics[width=\columnwidth]{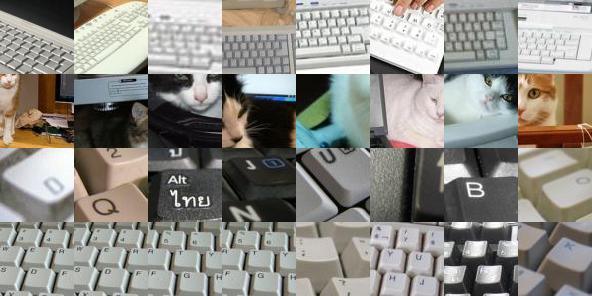}}
\caption{Concept based explanation for the class 'computer keyboard, keypad'. The second concept shows cats (in front of a monitor) which signifies a bias in the model/data, since the presence of cats should not be used to predict the presence of a computer keyboard}
\label{fig:keyboard}
\end{center}
\vskip -0.2in
\end{figure}

\begin{figure}[tb]
\vskip 0.2in
\begin{center}
\centerline{\includegraphics[width=\columnwidth]{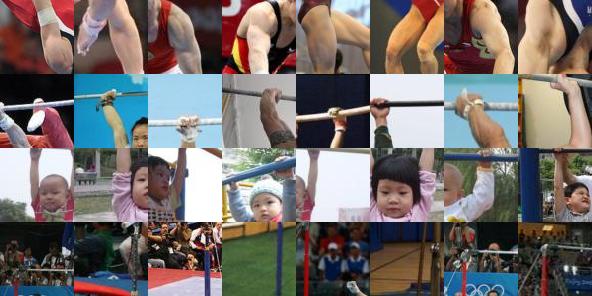}}
\caption{Concept based explanation for the class 'horizontal bar, high bar'}
\label{fig:bar}
\end{center}
\vskip -0.2in
\end{figure}

\begin{figure}[tb]
\vskip 0.2in
\begin{center}
\centerline{\includegraphics[width=\columnwidth]{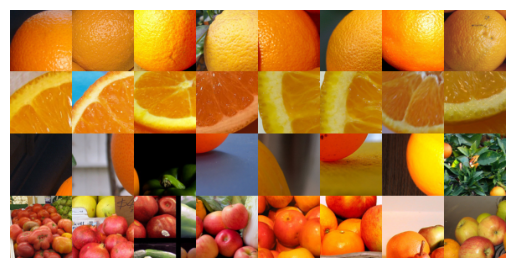}}
\caption{Visualization for the 4 NMF components for orange (fruit), using layer4.2. The model predicts 1.0 for orange for the four concepts (passing the first column into the model)}
\label{fig:orange}
\end{center}
\vskip -0.2in
\end{figure}

\begin{figure}[tb]
\vskip 0.2in
\begin{center}
\centerline{\includegraphics[width=\columnwidth]{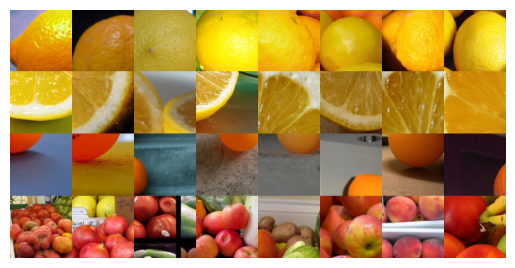}}
\caption{Visualization for the 4 NMF components for orange, using layer4.2 and excluding images where the model predicts orange. The model predicts 0.97 for orange for the four concepts}
\label{fig:orange_exclude_target}
\end{center}
\vskip -0.2in
\end{figure}

In this section we include some more examples for the ResNet50 model. For example, Figure \ref{fig:keyboard} shows a concept based explanation for why the model predicts 'computer keyboard, keypad'. Here the second concept shows cats in front of a monitor, meaning the model utilizes the presence of cats when predicting the class 'computer keyboard, keypad'. That can be problematic, because if is something is not a computer keyboard it does not become one if a cat is next to it and a computer keyboard is still one even if there is no cat next to it. Or Figure \ref{fig:bar} shows a concept based explanation for the class 'horizontal bar, high bar', but the first concept does not show any bar at all, only gymnasts. The second one shows hands on a bar and the third one shows small children hanging from a bar. Meaning the model will likely have problems correctly classifying a horizontal bar if there is no human on it.

Figure \ref{fig:orange_exclude_target} shows the four main concepts (excluding patches where the model predicts orange) for orange (fruit) for layer4.2 of the ResNet50 model from pytorch model zoo. Even taking the bottom two concepts (the edge of an orange ping pong ball and a bunch of fruits) combined gets predicted as orange with a confidence of 1.0. That points towards a problem in the model for this class (bunch of fruits plus edge of an orange ball gets predicted as orange). Also none of the four concepts actually shows orange as a plant, but only fruits or slices, which might also points towards a bias in the model.

\begin{figure}[tb]
\vskip 0.2in
\begin{center}
\centerline{\includegraphics[width=\columnwidth]{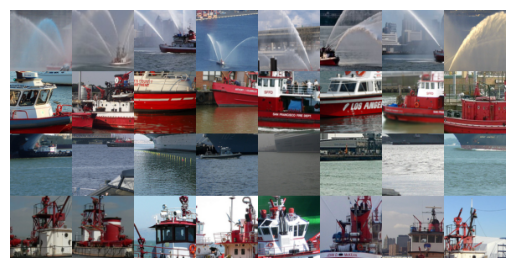}}
\caption{Visualization for the 4 NMF components for fireboat, using layer4.2. The model predicts 1.0 for fireboat for the four concepts (passing the first column into the model)}
\label{fig:goldfish_layer3}
\end{center}
\vskip -0.2in
\end{figure}

\begin{figure}[tb]
\vskip 0.2in
\begin{center}
\centerline{\includegraphics[width=\columnwidth]{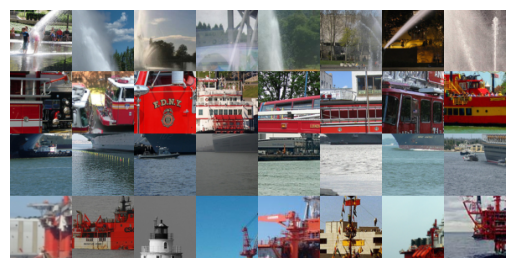}}
\caption{Visualization for the 4 NMF components for fireboat, using layer4.2 and excluding images where the model predicts fireboat. The model still predicts 1.0 for fireboat for the four concepts}
\label{fig:goldfish_layer3}
\end{center}
\vskip -0.2in
\end{figure}

\begin{figure}[tb]
\vskip 0.2in
\begin{center}
\centerline{\includegraphics[width=\columnwidth]{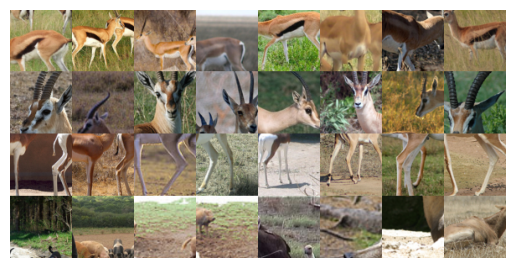}}
\caption{Visualization for the 4 NMF components for gazelle, using layer4.2. The model predicts 1.0 for gazelle for the four concepts}
\label{fig:goldfish_layer3}
\end{center}
\vskip -0.2in
\end{figure}

\begin{figure}[tb]
\vskip 0.2in
\begin{center}
\centerline{\includegraphics[width=\columnwidth]{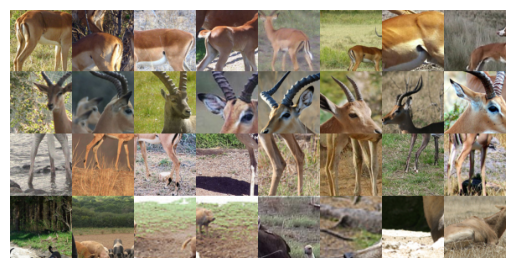}}
\caption{Visualization for the 4 NMF components for gazelle, using layer4.2 and excluding images where the model predicts gazelle. The model only predicts 0.05 for gazelle, with impala being the majority class with 0.95}
\label{fig:goldfish_layer3}
\end{center}
\vskip -0.2in
\end{figure}

\end{document}